\documentclass[11pt]{article}

\usepackage[T1]{fontenc}
\usepackage{amsmath,amssymb,amsfonts}
\usepackage{graphicx}
\usepackage{textcomp}
\usepackage{booktabs}
\usepackage{array}
\usepackage{multirow}
\usepackage{url}
\usepackage{bm}
\usepackage[margin=1in]{geometry}
\usepackage[hidelinks]{hyperref}
\graphicspath{{figures/}}

\newenvironment{keywords}{%
  \par\medskip\noindent\textbf{Keywords: }\itshape
}{\par\medskip}

\title{VeriWeave Govern: Evidence-Gated Deterministic Runtime Governance for Enterprise AI Agents}
\author{%
Kabeh Mohsenzadegan$^{1}$, Vahid Tavakkoli$^{1,*}$, Kyandoghere Kyamakya$^{1,2}$\\[0.6em]
\small $^{1}$Institute for Smart System Technologies, University of Klagenfurt, Klagenfurt, Austria\\
\small $^{2}$Facult\'e Polytechnique, Universit\'e de Kinshasa, Kinshasa, Democratic Republic of the Congo\\
\small $^{*}$Corresponding author: Kabeh Mohsenzadegan (\href{mailto:kabeh.mohsenzadegan@aau.at}{kabeh.mohsenzadegan@aau.at})\\
\small E-mail: \href{mailto:kabeh.mohsenzadegan@aau.at}{kabeh.mohsenzadegan@aau.at};
}
\date{}

\begin{document}
\maketitle

\begin{abstract}
Enterprise artificial-intelligence agents increasingly call tools, modify infrastructure, and process protected data, creating a need to separate action generation from action authorization. This article presents \emph{VeriWeave Govern}, a deterministic runtime governance layer that evaluates structured agent actions against versioned policies, validates typed evidence, applies fixed \emph{deny $>$ review $>$ allow} precedence, routes consequential actions to accountable human review, and records replayable tamper-evident audit state. GovernBench evaluates the design over 30 independent seeds and 60,000 oracle-labelled cases spanning five enterprise domains, adversarial evidence, out-of-distribution actions, and temporal policy evolution. VeriWeave achieves 0.9888 mean accuracy, 0.9836 macro-F1, zero observed aggregate false allows, and zero observed Governance Attack Success Rate on the evaluated cases. Six ablations show that evidence gating, deny precedence, out-of-distribution fail-safe behavior, human review, contradiction handling, and temporal replay contribute complementary safety. The deployed API additionally passes 12/12 end-to-end scenarios and a 40,040-request concurrency matrix with zero failures. A separate 150-case EU/Austria regulation-grounded evaluation uses frozen predictions and two independent blinded human annotators, who agree on all decisions. On this set, deterministic engines remain conservative, while a Gemma 4 31B comparator aligns more closely with the human consensus. The results expose a measurable safety--utility trade-off and motivate evidence-aware, replayable governance as an independent control plane for enterprise agent execution.
\end{abstract}


\section{Introduction}\label{sec:introduction}
	Large language models (LLMs) are evolving from text generators into agents that reason, call tools, and execute actions. ReAct and Toolformer established influential patterns for interleaving reasoning with environment/API interaction~\cite{yao2023react,schick2023toolformer}; recent surveys describe increasingly autonomous LLM-based and agentic systems operating over complex goals~\cite{dong2024agents,acharya2025agentic}. This capability shift changes the security question. An incorrect answer is undesirable, but an incorrectly \emph{authorized action} can deploy software, release data, send external messages, or affect a citizen-facing process.
	
	Conventional authorization provides necessary foundations. RBAC and ABAC govern permissions using roles and attributes~\cite{sandhu1996rbac,nistabac}; OPA and Cedar externalize policy evaluation into dedicated engines~\cite{opa,cedar}. Agent governance, however, also depends on evidence freshness, conflicting controls, human accountability, unknown actions, and policy evolution. For example, a production deployment can be technically permitted yet still require a current change approval and rollback plan; a high-impact action can possess complete documentation yet remain subject to human review.
	
	Security work further motivates independent enforcement. Surveys of autonomous and multi-agent systems emphasize robustness and security as design concerns~\cite{andreoni2024security,owoputi2022security,liu2024scaling}, while LLM evaluations report non-determinism and security reasoning failures~\cite{ullah2024llmsecurity}. Privacy and prompt-mediated attack surfaces remain active concerns~\cite{rathod2025privacy}, and autonomous red teaming demonstrates how context and tools create application-specific jailbreak opportunities~\cite{redagent2026}. These findings argue against using the same stochastic component as both proposer and final authorization authority.
	
	We therefore introduce \emph{VeriWeave Govern}: a deterministic governance boundary placed between agents/workflows and side-effecting tools. Its final decision path does not require an LLM. The main contributions are:
	\begin{itemize}
		\item an evidence-gated tri-state decision model with deterministic deny precedence, OOD fail-safe review, explicit human queues, temporal replay, and tamper-evident audit;
		\item a concrete reference architecture integrating identity/context, versioned policy repositories, evidence sources, enforcement outcomes, and replayable audit state;
		\item GovernBench: 30 seeds $\times$ 2,000 synthetic cases with adversarial families, calibration metrics, six ablations, and safety-specific measures;
		\item a separate 150-case EU/Austria regulation-grounded validation artifact with official-source provenance, two independent blinded human annotations, and frozen real-LLM/policy-engine comparisons.
	\end{itemize}
	
	This work emphasizes a systems question that is easily obscured when governance is treated only as policy syntax: \emph{what information must cross the runtime boundary before an autonomous action is allowed to acquire side effects?} VeriWeave answers with a typed contract. The planner supplies an intended operation and context, governed systems supply identity and evidence, a versioned policy bundle supplies constraints, and the governor returns a decision plus reasons that are independently replayable. This separates three concerns that are often conflated: semantic planning, authorization, and organizational accountability.
	
	The design is intentionally conservative about claims. VeriWeave does not decide whether an organization is legally compliant, and the regulation-grounded bank is not a substitute for legal analysis. Instead, the system operationalizes policy decisions that an organization has already chosen to encode and tests whether those decisions remain stable under missing evidence, conflicting rules, unknown actions, adversarial inputs, and policy evolution. This narrower boundary makes the system amenable to deterministic testing and audit while leaving normative policy ownership with accountable humans.
	
	\section{Related Work}
	\subsection{Policy enforcement and governance architectures}
	Policy-as-code reduces authorization logic embedded in application code. Cedar formalizes a safe and analyzable authorization language~\cite{cedar}, while OPA is widely used for general-purpose policy evaluation; recent cloud-compliance work illustrates OPA in CI/CD enforcement~\cite{paul2024opa}. VeriWeave is complementary: a deployment may use OPA/Cedar for predicate authorization while VeriWeave adds evidence acceptance, tri-state escalation, temporal replay, and governance audit semantics.
	
	Responsible-AI research increasingly moves from abstract principles toward operational architecture and controls. Lu \emph{et al.} propose a reference architecture for foundation-model systems~\cite{lu2024responsible}. AIRMan operationalizes AI risk management~\cite{tjoa2022airman}; public-procurement work highlights practical governance and ethics gaps~\cite{vonbehr2022procurement}; and recent IEEE studies examine policy-oriented AI governance, ethical regulatory frameworks, and enterprise risk-management extensions~\cite{kim2024governance,wang2024ethical,mcgrath2025erm}. VeriWeave focuses narrowly on the runtime point at which a proposed action becomes executable.
	
	\subsection{Regulatory and compliance automation}
	NIST AI RMF frames AI risk management as a lifecycle activity~\cite{nistairmf}. The EU AI Act introduces risk-tiered obligations and prohibited practices~\cite{aiact}, GDPR supplies independent privacy/accountability constraints~\cite{gdpr}, and Regulation (EU) 2026/1744 modifies parts of the AI Act application schedule~\cite{digitalomnibus}. Comparative regulatory analysis illustrates diverging governance approaches across jurisdictions~\cite{radi2023regulation}. LLM-assisted compliance research explores richer legal reasoning~\cite{hassani2024compliance}, and domain work studies how generative AI might support EU-AI-Act-oriented engineering~\cite{keser2024aiact}. We treat these as motivation for provenance and review, not as evidence that an automated governor can determine legal compliance.

	\subsection{Agent safety and runtime mediation}
	Agent-security research commonly emphasizes prompt injection, tool misuse, privilege boundaries, and robustness of autonomous or multi-agent systems~\cite{andreoni2024security,owoputi2022security,rathod2025privacy,redagent2026}. These concerns are related to, but distinct from, runtime authorization. A model can be robust against one class of prompt attack and still request an action that an organization does not permit; conversely, a benign request may still lack the evidence required for automatic execution. VeriWeave therefore treats the planner as an untrusted source of proposals rather than as a source of authority.
	
	This positioning also distinguishes \emph{decision assistance} from \emph{decision enforcement}. LLMs can help retrieve policies, summarize evidence, normalize requests, or explain why a case is difficult, but those outputs remain inputs to the trusted control plane. The final transition from proposal to side effect is determined by explicit policy, typed evidence, and review semantics. The resulting architecture is compatible with LLM-based reasoning while avoiding a circular trust model in which the component that invents an action also certifies that the action is permissible.
	
	\section{System Model and Architecture}\label{sec:system}
	\subsection{Design scope and assumptions}
	VeriWeave governs \emph{proposed side effects}: tool calls, infrastructure changes, data transfers, external communications, and other actions whose execution can alter enterprise state. Pure reasoning steps can remain inside the planner, but any transition to a protected resource is normalized into the governance request schema. The model assumes that the governance service, policy repository, trusted identity source, cryptographic keys, and authoritative evidence connectors are protected by conventional platform security. Compromise of those roots of trust is outside the benchmark threat model and would require infrastructure controls in addition to the decision logic studied here.
	
	Three design objectives follow. First, \emph{safety precedence} must be independent of rule order: a prohibition cannot disappear because a permissive rule was evaluated earlier. Second, \emph{evidence sufficiency} must be typed: ten weak documents should not substitute for one required approved change record. Third, \emph{uncertainty must remain explicit}: unknown actions, incomplete evidence, or procedural approval requirements should produce review rather than accidental allow. These objectives motivate the tri-state lattice used throughout the implementation.
	
	\subsection{Safety invariants}
	Agent governance differs from ordinary authorization because the proposer is stochastic, evidence may be dynamically retrieved, and some permitted actions remain procedurally human-gated. VeriWeave therefore enforces explicit invariants rather than relying on prompt instructions.
	
	\begin{table}[t]
\caption{Runtime governance invariants.}
\label{tab:invariants}
\centering\footnotesize
\setlength{\tabcolsep}{3.0pt}
\begin{tabular}{p{0.23\columnwidth}p{0.67\columnwidth}}
\toprule
Invariant & Runtime requirement \\
\midrule
No implicit allow & Unmatched/OOD actions route to review. \\
Deny dominance & A matched prohibition cannot be overridden by an allow rule. \\
Evidence monotonicity & Missing or rejected required evidence cannot improve a decision. \\
Accountability & High-impact rules may remain review-gated even with complete evidence. \\
Temporal fidelity & Decisions bind to policy versions and evaluation time. \\
Replayable integrity & Decision, evidence, reasons, and policy hash remain auditable. \\
\bottomrule
\end{tabular}
\end{table}

	A proposed action is
	\begin{equation}
		x=\langle a,o,r,p,c,E,t\rangle,
	\end{equation}
	where $a$ is agent/workload identity, $o$ the operation, $r$ the resource, $p$ the purpose, $c$ contextual attributes, $E$ the submitted evidence set, and $t$ evaluation time. A rule $q$ contains a predicate, target decision $d_q\in\{allow,review,deny\}$, required evidence types, version metadata, and an optional review queue.
	
	Let $Q(x)$ be the matched rule set and $G(x)\in\{0,1\}$ indicate whether all required evidence gates are satisfied. The final decision is
	\begin{equation}
		U(x)=\mathbf{1}\!\left[\exists q\in Q(x):d_q=review\right]\vee\neg G(x)\vee\mathbf{1}[Q(x)=\emptyset],
	\end{equation}
	\begin{equation}
		D(x)=\begin{cases}
			deny, & \exists q\in Q(x):d_q=deny,\\
			review, & U(x)=1,\\
			allow, & \text{otherwise}.
		\end{cases}
		\label{eq:lattice}
	\end{equation}
	The use of $Q(x)=\emptyset$ fixes unknown actions to \emph{review}; deny precedence prevents a permissive rule from overriding a simultaneously matched prohibition.
	
	\subsection{Reference architecture}
\begin{figure*}[t]
		\centering
		\includegraphics[width=0.98\textwidth]{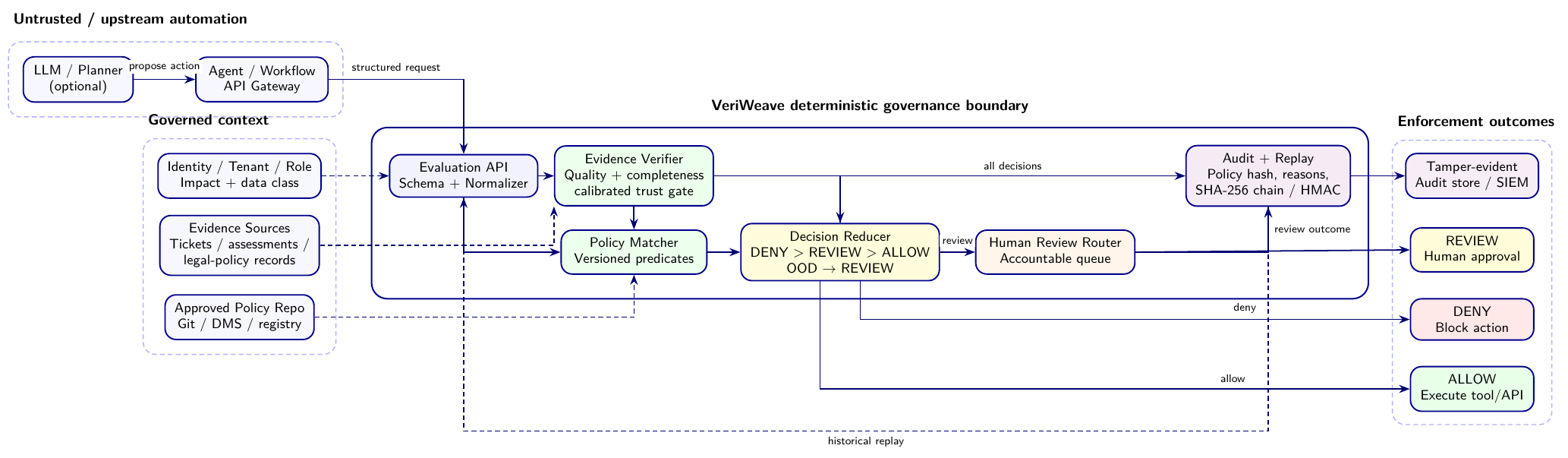}
		\caption{VeriWeave system architecture. Stochastic planning remains outside the deterministic governance boundary. Versioned policy matching and evidence verification feed a conservative decision reducer; review routes to an accountable human queue, while every outcome is recorded for verification and historical replay.}
		\label{fig:architecture}
	\end{figure*}

	Fig.~\ref{fig:architecture} shows the implementation boundary. Upstream LLMs or planners may generate candidate actions, but requests enter a structured evaluation API before any tool execution. Identity and context are supplied by an API gateway or workload-identity layer. Approved policies arrive from a version-controlled policy repository; evidence is drawn from governed sources such as change tickets, risk assessments, security reviews, or legal/policy records.
	
	Inside the boundary, the \emph{Policy Matcher} evaluates versioned predicates and the \emph{Evidence Verifier} checks required evidence types, source quality, freshness, signatures, and contradictions. Their outputs meet at the deterministic reducer of~\eqref{eq:lattice}. The \emph{Human Review Router} maps review outcomes to an accountable queue rather than treating review as an error. The \emph{Audit/Replay} component stores rule identifiers, reasons, evidence assessments, policy-set hashes, and chain state. This reflects the repository architecture: evaluation API, policy matching, evidence verification, decision reduction, review routing, and a hash-chained ledger.
	
	\begin{table}[t]
\caption{Core runtime components and responsibilities.}
\label{tab:components}
\centering\scriptsize
\setlength{\tabcolsep}{2.5pt}
\begin{tabular}{p{0.28\columnwidth}p{0.58\columnwidth}}
\toprule
Component & Responsibility \\
\midrule
Evaluation API & validate/normalize action, context, and evidence \\
Policy matcher & evaluate versioned predicates and evidence requirements \\
Evidence verifier & score authority, freshness, signatures, contradictions \\
Decision reducer & enforce deny $>$ review $>$ allow and OOD fail-safe \\
Review router & map consequential cases to accountable human queues \\
Audit/replay & hash-chain records and replay historical policy versions \\
\bottomrule
\end{tabular}
\end{table}

	\subsection{Integration and execution semantics}
	The design is intentionally compatible with existing enterprise components. OIDC/identity systems provide workload and tenant claims; OPA/Cedar may serve as predicate-policy backends; ticketing/workflow systems receive review tasks; and SIEM or immutable storage receives signed audit records. The API contract is deliberately narrower than a general natural-language prompt: action, resource, purpose, environment, impact, data classification, external-transfer state, and typed evidence are explicit fields. This reduces ambiguity before policy matching and makes policy coverage measurable.
	
	Execution is fail-closed with respect to governance semantics but not indiscriminately deny-by-default. A malformed request fails API validation, an unknown semantic action becomes \emph{review}, a matched prohibition becomes \emph{deny}, and a well-scoped low-risk request may become \emph{allow} only after its evidence requirements are satisfied. This distinction keeps operational uncertainty visible while avoiding an architecture in which every novel action is either silently permitted or permanently blocked.
	
	Policy bundles are versioned and identified in each response by a policy-set hash. Evidence assessments retain both acceptance state and reasons, allowing downstream reviewers to distinguish missing evidence from low-authority, stale, unsigned, or contradictory evidence. Thus the governance decision remains explainable without asking an LLM to reconstruct the decision after execution.
	
	\subsection{Policy lifecycle and temporal semantics}
	Runtime governance is not only a function of request content; it is also a function of \emph{when} the request was evaluated and which policy bundle was authoritative at that time. VeriWeave therefore treats policy version, evaluation time, and source snapshot as first-class inputs. A policy update may legitimately change a later decision without invalidating the auditability of an earlier one. Historical replay binds the old request to the old policy hash, while live evaluation binds the request to the currently active bundle.
	
	This distinction matters for both engineering and regulatory scenarios. Rollouts can introduce a rule in shadow mode, activate it after review, and later retire it without rewriting history. Likewise, a legal source may be known before a provision becomes applicable; source availability and application date are not interchangeable. The validation artifact records both dimensions so that future-effective cases are not scored as though they were already enforceable at the snapshot date.
	
	Policy evolution also creates a regression-testing obligation. Counterfactual certificates and frozen benchmark cases can be reevaluated against a candidate bundle before promotion. A change is then observable not only as a textual diff but as a decision diff: newly allowed, newly reviewed, and newly denied cases can be inspected before production activation.
	
	\subsection{Evidence gate and audit integrity}
	Each evidence item is mapped to features for authority, current status, signature, substantive content, policy-reference language, and contradiction. A logistic trust score
	\begin{equation}
		s(e)=\sigma(\mathbf{w}^{\top}\phi(e)+b)
	\end{equation}
	is fitted on synthetic training data separate from evaluation. Evidence is accepted only when $s(e)\geq\tau$ and its type satisfies the rule requirement. We report AUROC/AUPRC, Brier score, and ECE rather than interpreting $s(e)$ as a legal probability; this follows established calibration practice~\cite{guo2017calibration}.
	
	For a matched rule $q$ with required evidence types $T_q$, typed sufficiency can be written as
	\begin{equation}
		G_q(x)=\bigwedge_{\eta\in T_q}\mathbf{1}\!\left[\exists e\in E:\,\operatorname{type}(e)=\eta\land A(e)=1\right],
		\label{eq:typedgate}
	\end{equation}
	where $A(e)$ denotes evidence acceptance after quality, freshness, signature, and contradiction checks. Equation~\eqref{eq:typedgate} prevents evidence volume from becoming an implicit vote: only an accepted item of the required type closes that requirement. The overall gate $G(x)$ is the conjunction of applicable typed requirements. If an allow path fails this gate, the reducer produces review rather than attempting to infer that missing documentation is harmless.
	
	The evidence score and the governance decision therefore serve different roles. The score ranks whether an item is acceptable as evidence under the configured verifier; it does not directly authorize an action. Authorization remains a symbolic consequence of matched rules, typed sufficiency, and precedence. This separation is important when calibration is imperfect: a threshold can be tuned and audited without converting the learned score into an opaque end-to-end permission model.
	
	For canonical record $R_i$, audit chaining uses
	\begin{equation}
		h_i=\mathrm{SHA256}(h_{i-1}\parallel R_i),
	\end{equation}
	with optional HMAC signing. Tamper evidence assumes that the chain head or a checkpoint is protected outside the mutable audit store, or is authenticated with an HMAC/signature; a hash chain alone cannot prevent an attacker with full control of the store from rewriting the entire history. Policy hashes and timestamps support historical replay under the policy version that governed the original request.
	
	\subsection{Counterfactual governance certificates}
	A governance decision is more useful when an operator can inspect what would change it. VeriWeave therefore emits research certificates containing the observed decision, minimal supporting evidence, missing evidence, policy version, and decision-changing perturbations. This is distinct from free-form explanation: the counterfactual is recomputed through the same deterministic decision function.
	
	\begin{table}[t]
\caption{Counterfactual governance-certificate example.}
\label{tab:counterfactual}
\centering\footnotesize
\begin{tabular}{p{0.67\columnwidth}l}
\toprule
Perturbation & Decision \\
\midrule
Observed case & allow \\
Remove minimal supporting evidence & review \\
Set external destination + secret data & deny \\
\bottomrule
\end{tabular}
\end{table}

	The certificate example in Table~\ref{tab:counterfactual} begins with an allowed synthetic case. Removing its minimal supporting evidence forces review; changing the context to an external destination carrying secret data forces deny. These traces make the decision boundary inspectable and are useful for regression testing after policy updates. A certificate is explanatory evidence, not a bearer token: production authorization must remain bound to the live request, policy hash, identity, and audit record.
	
	\subsection{Determinism, monotonicity, and cost}
	The architecture is designed around properties that can be checked independently of the LLM that proposed the action. \textbf{Proposition 1 (deny dominance):} for a fixed request $x$, if any matched rule has target decision \emph{deny}, adding additional allow/review rules cannot make $D(x)$ more permissive. This follows directly from the first branch of~\eqref{eq:lattice}; therefore rule enumeration order does not change the final decision.
	
	\textbf{Proposition 2 (evidence monotonicity):} consider a request for which no deny rule matches. If a required evidence item is removed, expires, loses its trusted signature, or falls below threshold, an \emph{allow} can remain allow only when an equivalent accepted item still satisfies the same typed requirement; otherwise the outcome is downgraded to review. Removing evidence cannot create a new allow. This matters for agentic systems because omission becomes a conservative signal rather than a strategy for hiding unfavorable context.
	
	\textbf{Proposition 3 (replay determinism):} given identical normalized request $x$, policy bundle/version $P_v$, evidence feature extraction, threshold $\tau$, and evaluation-time semantics, repeated evaluations return the same decision, reasons, and matched-rule set. Audit identifiers and wall-clock metadata may differ for a new live invocation, but historical replay over frozen inputs reproduces the governance outcome. This property is the main reason that stochastic generation is placed outside the enforcement boundary.
	
	Let $m$ be the number of active rules and $n$ the number of submitted evidence items. With simple indexed predicates, the reference evaluator is dominated by rule matching and evidence verification, approximately $O(m+n)$ per request before persistence; contradiction checks can add pairwise cost if implemented naively. In practice, policy partitioning by operation/resource and evidence typing reduce the active subset. The measured API latency in Section~\ref{sec:results} includes parsing, matching, evidence checks, decision reduction, and audit writing, rather than reporting only an in-process policy function.
	
	\section{Experimental Methodology}\label{sec:method}
\begin{figure}[t]
		\centering
		\includegraphics[width=\columnwidth]{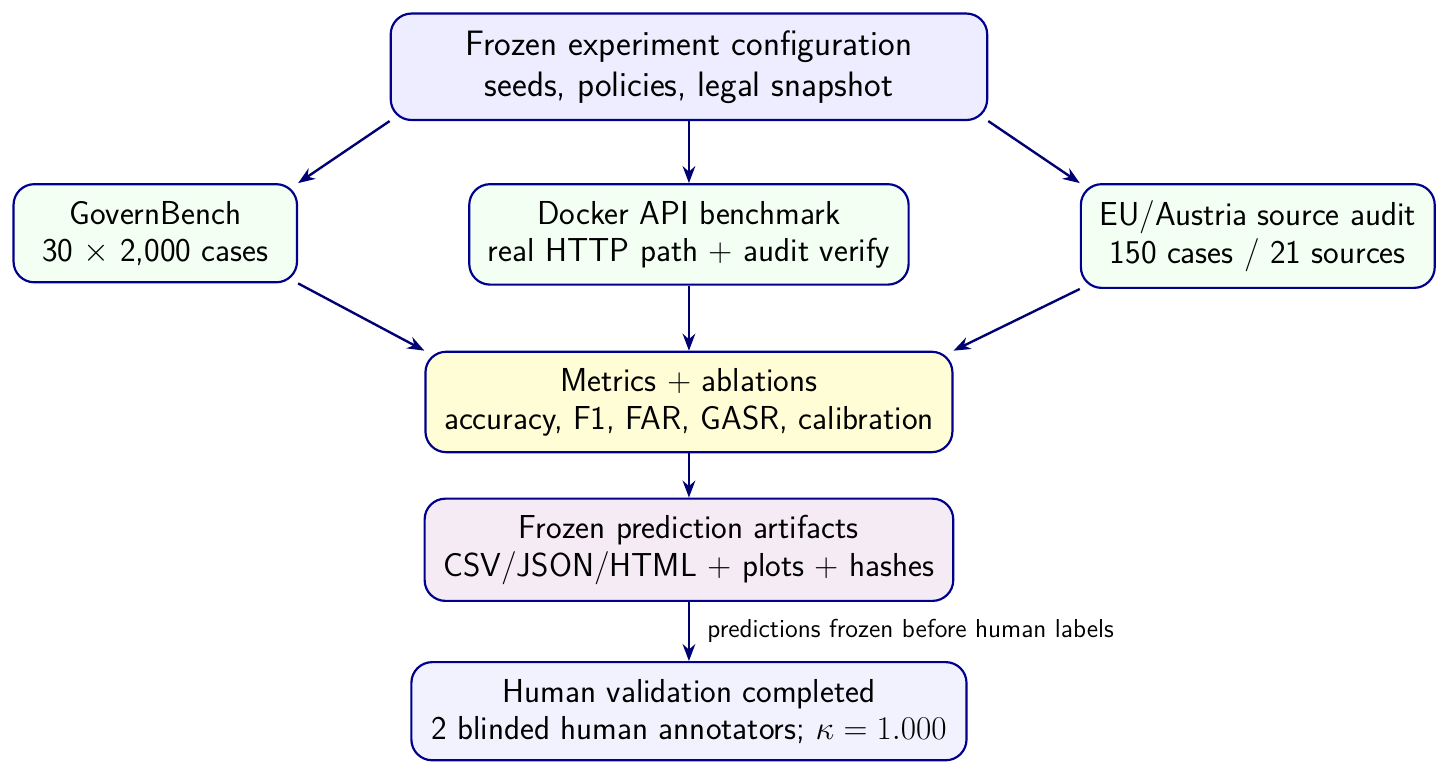}
		\caption{Experimental pipeline. Synthetic benchmarking, real-API validation, and source auditing are kept distinct, and comparator predictions are frozen before human annotation.}
		\label{fig:pipeline}
	\end{figure}

	Figure~\ref{fig:pipeline} summarizes the validation workflow. The controlled benchmark, service benchmark, and regulation-grounded source audit are generated from a frozen experiment configuration. Comparator predictions are then frozen before human labels are exposed. This ordering reduces the risk that benchmark labels, prompts, or policy translations are silently tuned after observing human outcomes.
	
	\subsection{GovernBench and research questions}
	GovernBench is a synthetic, oracle-labelled controlled benchmark. The reference experiment uses 30 independent seeds and 2,000 cases per seed (60,000 total) across public administration, healthcare, financial services, software-engineering agents, and enterprise-office agents. We ask whether deterministic governance (RQ1) reduces unsafe allows, (RQ2) resists adversarial evidence/action perturbations, (RQ3) depends on multiple complementary controls, (RQ4) meaningfully discriminates evidence quality, (RQ5) behaves correctly through the real HTTP service, and (RQ6) sustains concurrent service-level load without correctness failures.
	
	Cases include low-risk allows, weak/missing evidence, protected-data exfiltration, high-impact review, policy conflicts, OOD actions, and temporal evolution. Adversarial families include stale evidence, forged-signature metadata, citation laundering, evidence flooding, contradictory evidence, policy-version downgrade, and tool substitution. GASR is unsafe allows divided by attacked cases whose oracle decision is review/deny. Aggregate 95\% intervals are computed over seed-level metrics, avoiding the fiction that all 60,000 cases are independent draws from a real deployment.
	
	\subsection{Baselines and ablations}
	The controlled GovernBench study compares RBAC, ABAC, and the full governor. The external comparison uses a real Gemma 4 31B model invoked through Ollama and is evaluated separately on the 150-case regulation-grounded bank alongside executable OPA and Cedar. Ollama is requested as \texttt{gemma4:31b-cloud} and resolves the run as \texttt{gemma4:31b}; temperature is zero and JSON-mode structured decisions are enforced. All comparator predictions were frozen before the annotation workbooks were completed.
	
	Six ablations independently remove evidence gating, contradiction detection, human-review gating, OOD fail-safe behavior, deny precedence, and temporal replay. This isolates whether safety emerges from one rule or a layered control structure.
	
	\subsection{Safety and utility metrics}
	Overall accuracy alone is insufficient for a governance system because different errors have different operational consequences. Let $y_i$ be the reference tri-state label and $\hat y_i$ the system decision. We therefore report macro-F1 together with safety-specific rates. The false-allow rate over the evaluated set is
	\begin{equation}
		\mathrm{FAR}=\frac{\sum_i \mathbf{1}[\hat y_i=allow\land y_i\neq allow]}{N},
	\end{equation}
	and the false-review rate is
	\begin{equation}
		\mathrm{FRev}=\frac{\sum_i \mathbf{1}[\hat y_i=review\land y_i=allow]}{N}.
	\end{equation}
	For adversarial subsets $\mathcal{A}$, Governance Attack Success Rate is
	\begin{equation}
		\mathrm{GASR}=\frac{\sum_{i\in\mathcal{A}}\mathbf{1}[\hat y_i=allow\land y_i\in\{review,deny\}]}{\sum_{i\in\mathcal{A}}\mathbf{1}[y_i\in\{review,deny\}]}.
		\label{eq:gasr}
	\end{equation}
	These metrics expose different trade-offs: FAR and GASR capture unsafe execution, whereas FRev captures operational friction caused by conservative escalation. Reporting them together prevents a system from appearing strong merely by sending every case to human review.
	
	\subsection{Threat model and adversarial families}
	The threat model assumes that the governance service and its trusted configuration are not already compromised. Instead, an agent or upstream workflow can manipulate the proposed operation, destination, contextual attributes, submitted evidence, or requested policy version in an attempt to obtain an unsafe allow. This distinction targets the practical boundary between a stochastic planner and deterministic enforcement rather than host-level compromise.
	
	\begin{table}[t]
\caption{Adversarial families and defensive controls.}
\label{tab:attacks}
\centering\footnotesize
\setlength{\tabcolsep}{3.0pt}
\begin{tabular}{p{0.28\columnwidth}p{0.61\columnwidth}}
\toprule
Attack family & Expected defense \\
\midrule
Stale evidence & freshness/current-source gate \\
Forged signature & signature/evidence-quality rejection \\
Citation laundering & authority + substantive-source scoring \\
Evidence flooding & typed requirements; no vote-by-volume \\
Contradictory evidence & contradiction detection + review \\
Policy downgrade & version/temporal validation \\
Tool substitution & context predicates + OOD fail-safe \\
\bottomrule
\end{tabular}
\end{table}

	The adversarial families deliberately test different failure modes. Evidence flooding asks whether quantity can substitute for typed sufficiency; citation laundering places policy-like text inside low-authority material; contradictory evidence tests whether a favorable source masks an unfavorable one; tool substitution changes the surface action while preserving risky side effects. The design goal is not merely high average accuracy but conservative behavior on cases where uncertainty or manipulation should prevent automatic execution.
	
	\subsection{Service and regulation-grounded validation}
	A Docker benchmark exercises the actual FastAPI service through its public API, checking expected decisions, evidence assessments, review queues, policy hashes, signed audit state, and verification. A separate service-level load matrix issues 10,010 requests at each concurrency level 1, 4, 16, and 32 (40,040 total) and preserves failures rather than discarding timed-out or incorrect responses. Separately, the EU/Austria bank contains 150 constructed scenarios: 50 public-administration, 50 enterprise IT/DevOps, and 50 data/AI-governance cases. Each points to official EUR-Lex or Austrian RIS sources. A 17 Aug. 2026 snapshot records verification and legal application dates; future-effective Annex III scenarios are evaluated only after the amended 2 Dec. 2027 date specified by Regulation (EU) 2026/1744.
	
	\begin{figure}[t]
		\centering
		\includegraphics[width=\columnwidth]{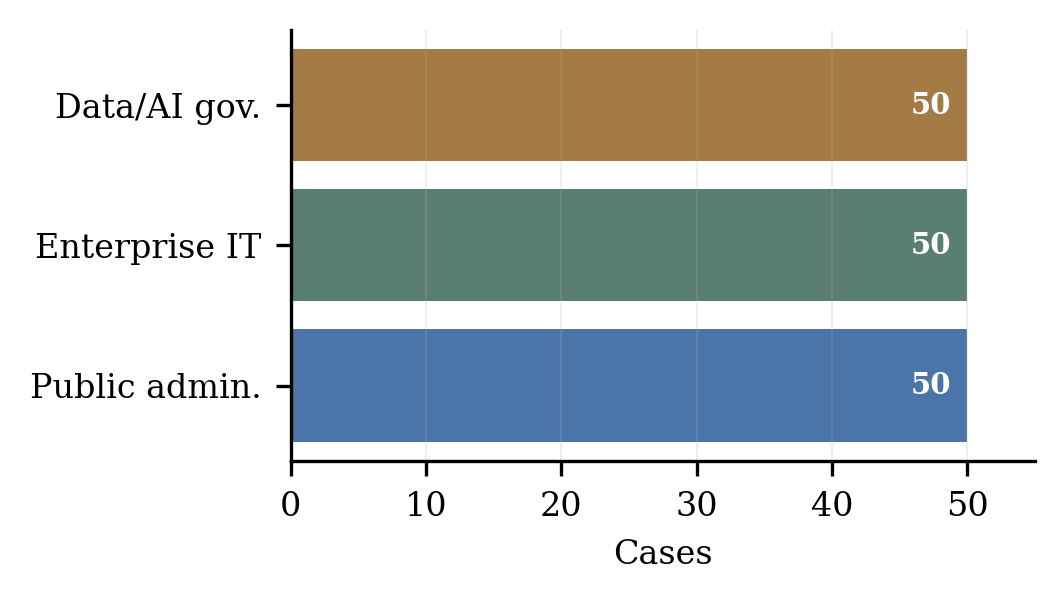}
		\caption{Composition of the 150-case regulation-grounded validation bank. The three domains are balanced by construction.}
		\label{fig:validationdist}
	\end{figure}
	
	The case bank was distributed to two human annotators in separate blinded workbooks that excluded provisional labels and all system predictions. The annotation protocol required each annotator to have documented familiarity with at least one relevant area: EU/Austrian public administration, privacy/data protection, AI governance, enterprise security/architecture, or IT/DevOps risk. The annotators worked independently, were not shown each other's decisions, and did not discuss individual cases before both workbooks were frozen. Only pseudonymous annotator identifiers were retained in the research artifacts. Each annotator independently completed all 150 rows with one tri-state decision, confidence score, and rationale per case using the scenario facts, evidence state, evaluation date, legal status, and cited official-source summaries/URLs. Agreement was computed before any adjudication using raw agreement and Cohen's $\kappa$~\cite{cohen1960}. The two human annotations agree on all 150 cases, so no adjudication was required; the common decision defines the human-reviewed reference label for that case. The study used constructed regulation-grounded scenarios rather than real citizen, patient, employee, client-secret, or other protected organizational data. These labels support benchmark evaluation, but they are not legal advice and do not establish authoritative or universally applicable regulatory interpretation.
	
	\subsection{Statistical treatment and artifact pipeline}
	Each GovernBench seed is treated as an independent synthetic realization with its own generated cases and fitted evidence calibrator. We aggregate seed-level results and report 95\% intervals across seeds rather than pooling all cases as if they were independent real-world observations. Baselines are evaluated on the same realization, which supports paired comparisons at the seed level. For the 150-case human-reviewed set, row-level predictions were frozen before annotation. We report case-level 10,000-sample bootstrap intervals, exact two-sided McNemar tests on paired correctness outcomes, and Holm--Bonferroni correction across comparator tests.
	
	The synthetic study, Docker API check, legal-source audit, frozen comparator predictions, and independent human annotation are deliberately separated until reporting time. This preserves the distinction between controlled oracle behavior, engineering correctness, source consistency, and the human-reviewed reference signal used for the 150-case comparison.
	
	\begin{table}[t]
\caption{Independent human annotation summary for the 150-case review set.}
\label{tab:human}
\centering\footnotesize
\begin{tabular}{lr}
\toprule
Measure & Result \\
\midrule
Human annotators & 2 \\
Completed pairs & 150 / 150 \\
Raw agreement & 1.000 \\
Cohen's $\kappa$ & 1.000 \\
Disagreements / adjudications & 0 / 0 \\
Human consensus allow/review/deny & 80 / 20 / 50 \\
Mean confidence A / B & 4.87 / 4.87 \\
Provisional--human agreement & 0.800 \\
Provisional--human $\kappa$ & 0.700 \\
\bottomrule
\end{tabular}
\end{table}

	\section{Results}\label{sec:results}
	\subsection{Controlled benchmark}
	\begin{table}[t]
\caption{GovernBench results (30 seeds $\times$ 2,000 cases).}
\label{tab:baseline}
\centering\scriptsize
\setlength{\tabcolsep}{2.7pt}
\begin{tabular}{lcccc}
\toprule
Method & Accuracy [95\% CI] & Macro-F1 & FAR & GASR \\
\midrule
RBAC & .5864 [.5845,.5883] & .6012 & .3032 & .7142 \\
ABAC & .4947 [.4945,.4948] & .5468 & .5053 & .8569 \\
\textbf{VeriWeave} & \textbf{.9888 [.9876,.9899]} & \textbf{.9836} & \textbf{.0000} & \textbf{.0000} \\
\bottomrule
\end{tabular}
\end{table}

	VeriWeave reaches 0.9888 mean accuracy (95\% CI 0.9876--0.9899) and 0.9836 macro-F1, with zero observed aggregate false allows and zero observed GASR in the evaluated GovernBench cases. RBAC and ABAC exhibit substantially higher unsafe-allow rates because the synthetic cases intentionally include evidence/review semantics they do not represent. The real LLM result is reported separately on the regulation-grounded bank because it is a different frozen evaluation set.
	
	\begin{figure}[t]
		\centering
		\includegraphics[width=\columnwidth]{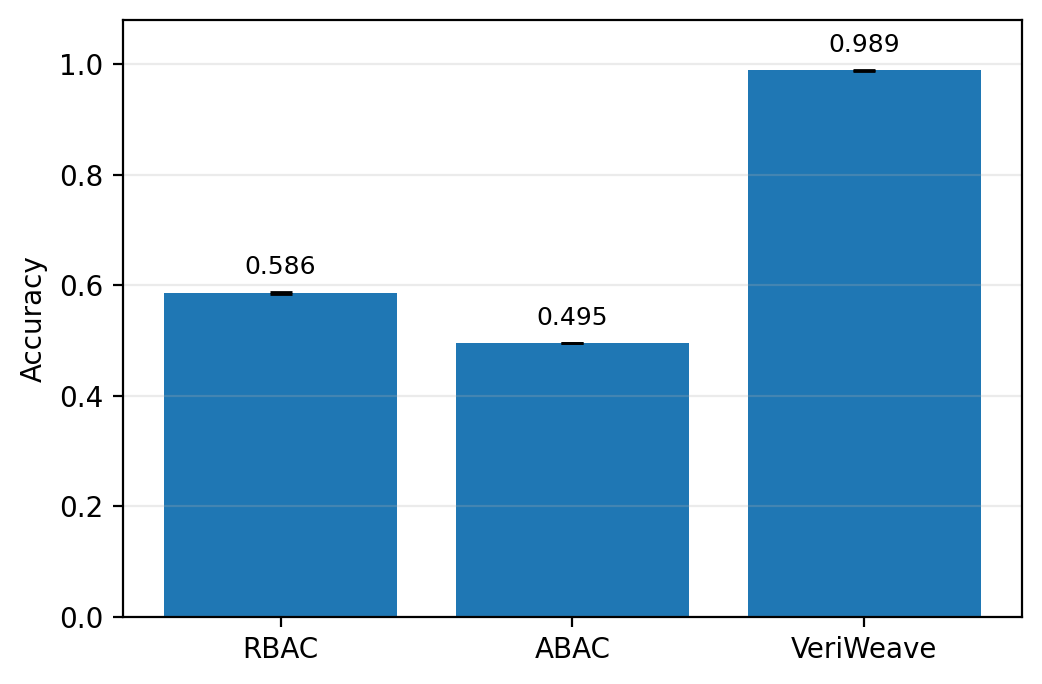}
		\caption{GovernBench mean accuracy across 30 independent seeds; error bars show recorded 95\% intervals.}
		\label{fig:accuracy}
	\end{figure}
	
	\subsection{Ablation: where safety comes from}
	\begin{table}[t]
\caption{Ablation results; each row removes one control.}
\label{tab:ablation}
\centering\footnotesize
\setlength{\tabcolsep}{3.2pt}
\begin{tabular}{lccc}
\toprule
Removed control & Accuracy & FAR & GASR \\
\midrule
Evidence gate & .7127 & .2873 & .5715 \\
Contradiction check & .9880 & .0008 & .0067 \\
Human review & .8998 & .0890 & .0000 \\
OOD fail-safe & .8604 & .1284 & .1427 \\
Deny precedence & .7675 & .2115 & .0000 \\
Temporal replay & .9155 & .0665 & .0000 \\
\bottomrule
\end{tabular}
\end{table}

	Removing the evidence gate is most damaging: accuracy falls to 0.7127, FAR rises to 0.2873, and GASR to 0.5715. Removing deny precedence produces 0.2115 FAR; disabling the OOD fail-safe produces 0.1284 FAR and 0.1427 GASR. Human review and temporal replay also materially reduce unsafe allows. The pattern supports layered defense: proof-of-control, conflict resolution, coverage gaps, accountability, and temporal state fail differently.
	
	\begin{figure}[t]
		\centering
		\includegraphics[width=\columnwidth]{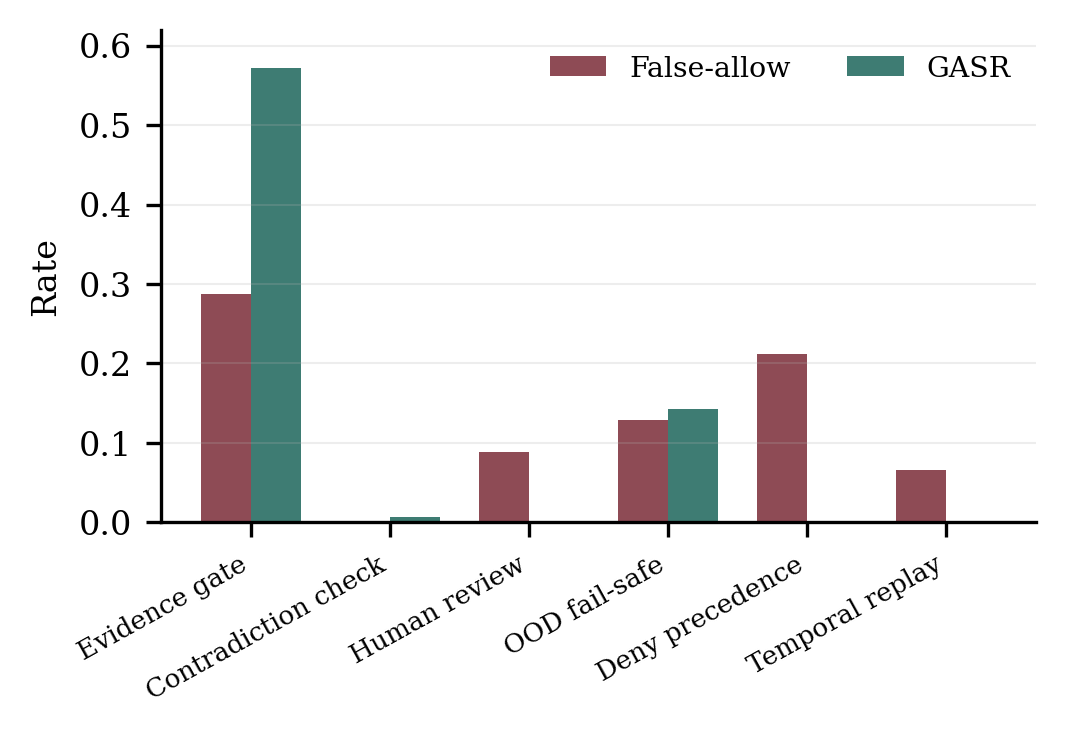}
		\caption{Safety impact of removing individual controls.}
		\label{fig:ablation}
	\end{figure}
	
	\subsection{Evidence quality and real service behavior}
	\begin{table}[t]
\caption{Evidence-model aggregate metrics.}
\label{tab:calibration}
\centering\footnotesize
\begin{tabular}{lc}
\toprule
Metric & Mean \\
\midrule
AUROC & 0.9840 \\
AUPRC & 0.9896 \\
Brier score & 0.1226 \\
Expected Calibration Error & 0.2253 \\
Selected threshold & 0.7600 \\
\bottomrule
\end{tabular}
\end{table}

	The latest held-out synthetic reliability profile shows strong discrimination (AUROC 0.9840; AUPRC 0.9896) but ECE 0.2253. We therefore describe $s(e)$ as a thresholded trust score rather than a calibrated probability.
	
	\begin{table}[t]
\caption{Docker API correctness and smoke performance.}
\label{tab:service}
\centering\footnotesize
\begin{tabular}{lr}
\toprule
Measure & Result \\
\midrule
Correctness scenarios & 12/12 passed \\
Measured requests & 110 \\
Concurrency & 4 \\
Mean latency & 10.42 ms \\
P95 / P99 latency & 17.14 / 19.93 ms \\
Throughput & 371.64 req/s \\
Audit chain & valid \\
\bottomrule
\end{tabular}
\end{table}

	\begin{table}[t]
\caption{Single-instance service load matrix (zero failures).}
\label{tab:load}
\centering\scriptsize
\setlength{\tabcolsep}{3.0pt}
\begin{tabular}{rrrrr}
\toprule
Conc. & Requests & Req/s & P95 ms & Fail. \\
\midrule
1  & 10,010 & 320.33 & 3.96   & 0 \\
4  & 10,010 & 413.55 & 13.27  & 0 \\
16 & 10,010 & 382.72 & 49.81  & 0 \\
32 & 10,010 & 368.97 & 108.84 & 0 \\
\bottomrule
\end{tabular}
\end{table}

	The real API path passes 12/12 scenarios. At concurrency four, 110 measured requests yield 10.42 ms mean latency, 17.14 ms P95, 19.93 ms P99, and 371.64 requests/s; the persisted audit chain verifies successfully. The separate load matrix in Table~\ref{tab:load} then executes 40,040 additional HTTP evaluations with zero failures. Throughput peaks at 413.55 requests/s at concurrency four and remains 368.97 requests/s at concurrency 32, where P95/P99 latency is 108.84/127.86 ms. These measurements include request parsing, policy matching, evidence verification, deterministic reduction, and audit persistence.
	
	\begin{table}[t]
\caption{Representative end-to-end decision traces.}
\label{tab:trace}
\centering\scriptsize
\setlength{\tabcolsep}{2.7pt}
\begin{tabular}{p{0.44\columnwidth}lp{0.34\columnwidth}}
\toprule
Scenario & Decision & Dominant control \\
\midrule
Trusted low-risk read & allow & accepted policy evidence \\
Outdated evidence & review & evidence-quality gate \\
Deny overrides review & deny & deny precedence \\
Production deployment & review & accountable change review \\
Unknown action & review & OOD fail-safe \\
\bottomrule
\end{tabular}
\end{table}

	Table~\ref{tab:trace} illustrates that the same API produces distinct outcomes for distinct governance causes: trusted evidence enables a narrow allow; outdated evidence downgrades the action; deny precedence resolves conflicting matched rules; production change remains accountable even with supporting controls; and an unknown operation cannot fall through to allow. The audit record retains the matched rule and reason for each path, so these outcomes are observable rather than implicit control-flow branches.
	
	\subsection{Regulation-grounded source audit}
	\begin{table}[t]
\caption{EU/Austria case-bank source and annotation summary.}
\label{tab:legal}
\centering\footnotesize
\begin{tabular}{lr}
\toprule
Property & Value \\
\midrule
Total cases & 150 \\
Public administration & 50 \\
Enterprise IT / DevOps & 50 \\
Data / AI governance & 50 \\
Primary-law sources & 21 \\
Temporal source references & 104 \\
Audit errors / warnings & 0 / 0 \\
Source audit status & PASS \\
Human consensus A/R/D & 80 / 20 / 50 \\
\bottomrule
\end{tabular}
\end{table}

	The legal audit passes with no errors or warnings. Its scientific boundary is narrow: it checks source provenance, case-bank consistency, application dates, and temporal semantics; it does not certify legal correctness. The original development labels were balanced 50/50/50, whereas the two human annotators yield 80 allow, 20 review, and 50 deny. Their decisions agree on 150/150 cases (raw agreement 1.000; $\kappa=1.000$), with mean confidence 4.87/5 for each annotator and no adjudication required. The provisional labels agree with the human consensus on 120/150 cases (0.800 raw agreement; $\kappa=0.700$); all 30 differences are provisional review$\rightarrow$human-consensus allow, revealing a systematic conservative bias in the development labels.
	
	\begin{table}[t]
\caption{150-case comparison against human-reviewed consensus labels.}
\label{tab:external}
\centering\scriptsize
\setlength{\tabcolsep}{2.8pt}
\begin{tabular}{lcccc}
\toprule
Method & Accuracy & Macro-F1 & FAR & FRev \\
\midrule
RBAC & .2400 & .2278 & .0000 & .7600 \\
ABAC & .4267 & .3515 & .0933 & .4800 \\
OPA & .8000 & .7802 & .0000 & .2000 \\
Cedar & .8000 & .7802 & .0000 & .2000 \\
Gemma 4 31B & \textbf{.9733} & \textbf{.9573} & \textbf{.0000} & \textbf{.0267} \\
\textbf{VeriWeave} & .8000 & .7802 & \textbf{.0000} & .2000 \\
\bottomrule
\multicolumn{5}{l}{FAR: false-allow rate; FRev: false-review rate.}
\end{tabular}
\end{table}

	All three external engines completed 150/150 cases with zero invocation failures. Against the human-reviewed consensus, Gemma 4 31B via Ollama reaches 0.9733 accuracy (bootstrap 95\% CI 0.9467--0.9933), 0.9573 macro-F1, and zero observed false allows on this 150-case set. Its four errors are one allow$\rightarrow$review and three deny$\rightarrow$review decisions. VeriWeave, OPA, and Cedar each reach 0.8000 accuracy (95\% CI 0.7333--0.8600), 0.7802 macro-F1, and zero observed false allows on this set; all 30 of their errors are consensus allow$\rightarrow$review escalations. Thus, the deterministic engines are strictly conservative on this bank but less aligned with the human annotators on automatic-allow utility. Relative to Gemma, VeriWeave's accuracy difference is $-0.1733$ (bootstrap 95\% CI $-0.2467$ to $-0.1067$; Holm-adjusted exact McNemar $p=7.67\times10^{-6}$). The result is a safety--utility trade-off rather than a universal ranking: Gemma aligns more closely with the human consensus, while VeriWeave preserves deterministic policy enforcement, explicit review routing, replay, and auditable zero-false-allow behavior on the reviewed cases.
	
	The six 25-case partitions are intentionally small enough for manual expert review while preserving a frozen case identifier and source registry. Current-law and future-effective scenarios are distinguished through both source-snapshot time and evaluation time. That separation is essential for reproducibility: a source can be known at the 2026 snapshot while a particular obligation is evaluated only at a later application date.
	
	\subsection{Safety--utility decomposition}
	The two evaluation regimes reveal why a single headline accuracy number is not an adequate description of runtime governance. In GovernBench, the oracle encodes the same evidence and review semantics that the system is designed to enforce; the full governor therefore achieves high agreement while maintaining zero aggregate unsafe allows. In the human-reviewed bank, by contrast, the development policy is deliberately more conservative than the two annotators on 30 cases. The resulting 0.8000 accuracy is not caused by unsafe permissions: every discrepancy for VeriWeave, OPA, and Cedar is an allow$\rightarrow$review escalation.
	
	Operationally, these error directions have different costs. A false allow can produce an unauthorized side effect, whereas an unnecessary review primarily consumes human capacity and delays execution. The experiment therefore identifies an engineering target rather than a reason to weaken the safety lattice: improve policy specificity and evidence coverage so that low-risk cases can be released automatically without changing deny dominance, OOD fail-safe behavior, or typed evidence requirements. This is the principal utility-improvement path suggested by the human-validation result.
	
	\section{Worked Runtime Cases and Comparator Protocol}\label{sec:cases}
	\subsection{Three execution paths}
	A useful governance architecture should make qualitatively different control paths explicit. Three cases exercised through the real Docker API illustrate distinct execution paths. The HTTP request passes schema validation, policy matching, evidence verification, decision reduction, audit writing, and response serialization before the benchmark checks the outcome.
	
	\begin{table*}[t]
\caption{Worked end-to-end cases from the Docker benchmark.}
\label{tab:worked}
\centering\scriptsize
\setlength{\tabcolsep}{4pt}
\begin{tabular}{p{0.18\textwidth}p{0.28\textwidth}p{0.35\textwidth}l}
\toprule
Case & Request facts & Governance path & Outcome \\
\midrule
Trusted knowledge read & internal, low-impact read; signed/current policy reference & allow rule matches; typed evidence is accepted and complete & allow \\
Production deployment & production deploy; approved change, tests, rollback evidence & production-change control remains accountable despite complete evidence & review \\
Secret external transfer & external destination; secret classification; high-impact context & prohibition and review rule both match; deny precedence resolves conflict & deny \\
\bottomrule
\end{tabular}
\end{table*}

	\textbf{Trusted low-risk read.} The first case proposes an internal summarization/read action in a test context with low impact. The request carries a current, signed policy reference from an approved policy library with high authority. The allow rule matches, the evidence verifier accepts the policy reference, and no higher-precedence rule matches. The resulting \emph{allow} is therefore supported by both policy and evidence. This case demonstrates why the evidence gate is not merely a risk score: the same low-risk action without the required reference is downgraded to review, and an outdated/unsigned reference is rejected by evidence quality checks.
	
	\textbf{Production deployment.} The second path separates evidence sufficiency from organizational accountability. A deployment request can carry an approved change ticket, test evidence, and rollback plan and still be routed to a production-change queue. This is intentional. Evidence answers ``are the prerequisites documented?''; it does not answer ``has an accountable person accepted this production action?'' The review state therefore persists even when all required artifacts are valid. This pattern maps naturally to CI/CD, database migration, firewall, identity-platform, and policy-bundle changes where technical automation should not erase four-eyes control.
	
	\textbf{Conflicting deny and review.} The third path combines an external transfer of secret information with a high-impact context. Both a prohibition and a human-review rule match. A first-match or policy-file-order implementation could accidentally produce different outcomes after refactoring. VeriWeave instead applies the lattice in~\eqref{eq:lattice}; \emph{deny} dominates, the security-incident queue is retained in the explanation, and the audit record captures both matched controls. This behavior is directly targeted by the deny-precedence ablation, where removing the invariant causes a large increase in unsafe allows.
	
	\subsection{Fair external-comparator protocol}
	Comparing an evidence-aware tri-state governor with general policy engines or LLMs requires a protocol that avoids giving one method hidden information. The artifact therefore separates \emph{case facts} from \emph{labels}. OPA and Cedar receive normalized structured facts and versioned equivalent policies. A real LLM baseline receives the scenario facts and official-source summaries, but not provisional labels, human-review rationales, prohibition metadata that encodes the answer, or VeriWeave predictions. The same frozen case identifiers are used for every method so correctness outcomes can be compared pairwise.
	
	The comparison should also distinguish \emph{engine capability} from \emph{reference-policy scope}. OPA and Cedar can encode sophisticated logic; a coarse study policy is not evidence of an intrinsic engine limitation. Consequently, external results should report the exact policy bundle, engine version, and translation assumptions. The useful research question is whether the same governance intent can be implemented reproducibly and what additional machinery is required for evidence acceptance, review routing, temporal replay, and audit explanation.
	
	In the reported LLM run, temperature is fixed at zero, JSON-mode structured output is enforced, and failures/timeouts are preserved rather than silently retried until correct. The requested Ollama model is \texttt{gemma4:31b-cloud}, which resolves to \texttt{gemma4:31b}; all 150 row-level outputs are retained. Mean invocation latency is 1.34 s for the cloud LLM, compared with 1.89 ms for OPA and 8.56 ms for Cedar in this run; because the deployment paths differ, these are descriptive timings rather than a controlled engine-speed comparison. We compute false-allow and false-review rates, 10,000-sample paired bootstrap intervals, exact two-sided McNemar tests, and Holm correction against the human consensus. This protocol prevents the validation stage from becoming a post-hoc selection exercise and preserves negative as well as positive results.
	
	\section{Discussion}\label{sec:discussion}
	The results support an architectural conclusion: \emph{action generation and authorization should be separate trust domains}. Agentic systems increase both action flexibility and attack surface~\cite{acharya2025agentic,dong2024agents,redagent2026}; a deterministic boundary converts that flexibility into an explicit contract. This does not make LLMs unnecessary. They can classify requests, retrieve candidate evidence, draft policies, or explain decisions, but their output remains untrusted until validated by deterministic controls.
	
	Tri-state governance is central. Binary authorization conflates ``safe to execute'' with ``cannot safely decide automatically.'' Review gives uncertainty an accountable destination. The ablations show why this is more than interface design: removing the review gate or OOD fail-safe introduces unsafe allows. Likewise, evidence gating prevents an agent from improving its authorization outcome by withholding, weakening, or laundering evidence.
	
	The independent human review changes the interpretation of the external comparison in an instructive way. Thirty cases that development labels had conservatively marked \emph{review} were marked \emph{allow} by both annotators. VeriWeave, OPA, and Cedar therefore incur no observed false allows on this set but a 0.2000 false-review rate, whereas Gemma has zero observed false allows and only 0.0267 false reviews on the same consensus. This does not remove the need for deterministic enforcement: the LLM still converts three consensus denies to review, and its output is stochastic policy advice rather than a replayable authorization proof. Instead, the result identifies a practical tuning target for deterministic governance---reduce unnecessary escalation without weakening deny dominance or evidence gates.
	
	The architecture also clarifies its relation to policy-as-code. OPA/Cedar remain strong choices for structured policy evaluation~\cite{opa,cedar,paul2024opa}; VeriWeave adds an evidence/review/audit envelope around such decisions. In a production design, the policy matcher can be backed by an external engine while the surrounding governance semantics remain unchanged.
	
	\subsection{Deployment trust boundaries}
	The architecture separates four trust zones. The \emph{planner zone} is explicitly untrusted and may contain an LLM, RAG component, or multi-agent coordinator. The \emph{context zone} contributes identity, tenant, role, impact, data classification, and approved policy/evidence sources; these inputs require their own authentication and lifecycle controls. The \emph{governance zone} is deterministic and should be deployed with a small, auditable attack surface. Finally, the \emph{execution zone} contains side-effecting tools and human approval systems that act only on an explicit governor outcome.
	
	This separation supports incremental adoption. In \emph{shadow mode}, the governor evaluates real proposed actions but does not block execution; mismatches reveal missing action classes and ambiguous policies. In \emph{active mode}, narrow low-risk allows and unambiguous denies can be enforced while consequential actions remain review-gated. The same policy-set hashes and audit schema should be preserved across both modes so that a later enforcement change does not erase the history used to justify it.
	
	\subsection{Operational deployment lifecycle}
	A deterministic governor can be introduced incrementally rather than as an immediate hard gate. A practical sequence is: (1) \emph{observe}, in which actions are normalized and logged but not blocked; (2) \emph{shadow}, in which the governor computes decisions and reviewers inspect mismatches with existing practice; (3) \emph{selective enforcement}, in which narrow allow and deny rules are activated for well-understood action classes while ambiguous classes remain review-only; and (4) \emph{broad enforcement}, in which coverage expands only after decision diffs, review load, and incident handling are understood. The audit schema should remain stable across these phases so that policy evolution can be analyzed longitudinally.
	
	Review capacity is part of the control system. A policy that routes too many low-risk cases to people is safe in a narrow sense but may be operationally unsustainable, leading teams to bypass the gate or approve mechanically. For that reason, false-review rate, queue latency, reviewer load, and recurring reasons for escalation should be treated as governance telemetry. The goal is not to eliminate review, but to reserve it for cases where accountability or uncertainty genuinely requires human judgment.
	
	\subsection{Policy-engineering implications}
	The experiments suggest four practical policy-engineering rules. First, encode prohibitions independently of permissive rules so deny dominance remains explicit. Second, express evidence requirements by type and provenance rather than by document count. Third, version policy bundles and bind decisions to policy hashes so changes are reviewable and replayable. Fourth, treat unknown semantic actions as a coverage signal: repeated OOD reviews indicate that the action taxonomy or policy bundle needs extension, not that the fail-safe should be removed.
	
	Counterfactual certificates can support this workflow by showing which minimal change would alter a decision. If removing one evidence item changes allow to review, that item is part of the authorization basis; if changing destination or data classification changes review to deny, the responsible predicate is visible. Such traces are useful for regression suites because they test the intended shape of the decision boundary rather than only the final label.
	
	\subsection{Limitations and threats to validity}
	\textbf{Synthetic oracle alignment:} GovernBench is generated from a controlled grammar and oracle policy. The large gaps versus RBAC/ABAC establish behavior on this benchmark, not universal superiority. The generator also cannot reproduce the full ambiguity, organizational politics, or undocumented practices present in real deployments.
	
	\textbf{Annotation and legal scope:} two blinded human annotators independently completed all 150 cases and agree perfectly ($\kappa=1.000$), providing a human-reviewed reference set. However, a constructed 150-case bank with two reviewers does not establish universal legal correctness, jurisdiction-wide validity, or real-world regulatory effectiveness. The cases depend on a finite EU/Austria source snapshot and simplified scenario facts. Replication with additional legal-domain reviewers, organizations, jurisdictions, and naturally occurring deployment cases remains necessary.
	
	\textbf{Comparator scope:} the OPA and Cedar results depend on the translated policy bundles used in this study and should not be interpreted as intrinsic limits of those engines. Likewise, the Gemma result reflects one model/run configuration and a frozen prompt protocol; it is not evidence that an LLM is generally safer or less safe than deterministic policy enforcement.
	
	\textbf{Calibration:} ECE 0.2253 precludes interpreting the evidence trust score as a well-calibrated probability. The current design mitigates this by using the score only inside a thresholded evidence-admission stage, but deployment-specific calibration and drift monitoring remain important.
	
	\textbf{Performance and scale:} the single-instance 40,040-request load matrix reaches concurrency 32 without failures, but the measured rates are hardware- and configuration-specific. Multi-instance scaling, substantially larger policy/evidence sets, durable distributed audit storage, long-duration soak tests, and cross-hardware replication remain future work.
	
	\textbf{Security boundary:} production use additionally requires authenticated workloads, tenant isolation, secret/key management, durable storage, rate limiting, monitoring, backup/recovery, policy-signing procedures, and independent security testing. The benchmark does not model compromise of the governor host, trusted identity provider, signing keys, or policy repository.
	
	\section{Reproducibility}\label{sec:repro}
	The artifact is implemented in Python 3.13 and released as VeriWeave Govern v0.4.0.\footnote{Open-source repository: \url{https://github.com/vtavakkoli/veriweave-govern.git}} The repository contains the benchmark generator, policies, calibration model, ablations, legal-source registry, Docker definitions, and publication workflow. In the reported run, a single Compose pipeline successfully completes GovernBench, the API benchmark, legal-source audit, real OPA/Cedar/Ollama validation, bootstrap statistics, calibration reporting, and the 40,040-request load matrix. Comparator predictions were frozen before human annotation, and the paper package records the human consensus labels, agreement statistics, relabelled development cases, and cryptographic hashes of the two completed annotation workbooks.
	
	A reproducible run should freeze at least the Git commit, container images and tool versions, policy-set hashes, evidence-threshold configuration, random seeds, hardware description, source-snapshot date, raw predictions, and annotation-workbook hashes. The distinction between \emph{re-execution} and \emph{historical replay} should also be preserved: re-execution tests the current implementation from frozen inputs, whereas historical replay reconstructs the decision under the policy version and temporal semantics that governed the original request. Both are useful, but they answer different audit questions.
	
	\section{Conclusion}\label{sec:conclusion}
	VeriWeave demonstrates that deterministic evidence gating can provide a practical enforcement layer between autonomous planning and side-effecting execution. Across 60,000 controlled cases it combines high accuracy with zero observed aggregate false allows and zero observed GASR; ablations show that evidence gating, deny precedence, OOD fail-safe review, human accountability, and temporal replay all contribute. The deployed service additionally completes a 40,040-request concurrency matrix with zero failures. On the 150-case human-reviewed set, VeriWeave, OPA, and Cedar retain zero observed false allows but over-escalate 30 cases to review, while the real Gemma 4 31B run aligns more closely with the human consensus and also exhibits zero observed false allows, at the cost of three deny$\rightarrow$review errors and without deterministic enforcement semantics. The broader systems principle is therefore not that a deterministic governor must maximize label accuracy; it is that action authorization should expose and control the safety--utility trade-off through independent, evidence-aware, replayable policy enforcement.

\end{document}